# Contextual Peano scan and fast image segmentation using hidden and evidential Markov chains[†]

**Clément Fernandes [1,2], Wojciech Pieczynski [2,*]**

**Abstract:** Transforming bi-dimensional set of images pixels into mono-dimensional sequences with Peano scan (PS) is a established technique enabling the use of hidden Markov chains (HMCs) for unsupervised image segmentation. Related Bayesian segmentation methods can compet with hidden Markov fields (HMFs) based ones, and are much faster. The PS been recently been extended to the contextual PS, and some initial experiments have shown the value of the associated HMC model, denoted as HMC-CPS, in image segmentation. Moreover, HMCs have been extended to hidden evidential Markov chains (HEMCs), which are capable of improving HMC-based Bayesian segmentation. In this study, we introduce a new HEMC-CPS model by simultaneously considering contextual PS and evidential HMC. We show its effectiveness for Bayesian maximum posterior mode (MPM) segmentation using synthetic and real images. Segmentation is performed in an unsupervised manner, with parameters being estimated using the stochastic expectation–maximization (SEM) method. The new HEMC-CPS model presents potential for the modeling and segmentation of more complex images, such as three-dimensional or multi-sensor multi-resolution images. Finally, the HMC-CPS and HEMC-CPS models are not limited to image segmentation and could be used for any kind of spatially correlated data.

## 1. Introduction

With the recent advancements in deep learning techniques, a considerable number of deep learning models have been designed and trained to perform unsupervised image segmentation. Indeed, from classical convolutional encoder–decoders [1] to recurrent networks [2], generative adversarial networks [3] and attention encoder–decoders [4], all these methods have produced spectacular results in various domains such as semantic segmentation, cell image segmentation and super-resolution reconstruction, outperforming the more traditional approaches. However, these models still require large training databases, even if the data are unlabeled. In this article, we focus on "agnostic" statistical methods, which only require the information present in the image for segmentation, and no additional sources. In this context, methods based on hidden Markov fields (HMFs) have been widely used since the pioneering research was performed [5], [6], [7]. HMFs provide satisfactory results in various situations, notably in image segmentation problems [8], [9], [10], [11], [12]. However, direct fast calculations are not tractable and iterative methods such as Gibbs or Metropolis sampling are required. These calculations significantly increase the computation time and make the related methods hard to apply in some situations. Utilizing the well-known and widely used hidden Markov chains (HMCs) [13], [14], [15], [16], [17] instead of HMFs is feasible; however, transforming a bi-dimensional set of pixels into a mono-dimensional sequence is not straightforward. For example, proceeding "line by line" gives an HMC in which pixels close each to other in the image may be far in the sequence. Using the Peano scan partially overcomes these difficulties and allows one to bring the quality of the results obtained with the chains closer to those obtained with the fields [18], [19]. The combination of the Peano scan and HMCs has been used to segment different types of images [20], [21], [22], [23], [24], [19]. Peano scans have also been used in more sophisticated models than HMCs; for example, hidden fuzzy Markov chains [25], [26], [19] or pairwise Markov chains [26]. Even if data obtained from images via the Peano scan have a complex structure and are obviously not Markovian, the different mentioned methods can provide interesting results, demonstrating the extraordinary robustness of HMCs and their extensions.

All the studies mentioned above show the value of the Peano scan in problems where the computation time is important; indeed, thanks to direct, recursive and exact computations, HMC-based unsupervised segmentations are incomparably faster than HMF-based ones.

We recently extended the Peano scan (PS) to the following "contextual" Peano scan (CPS) [27]. Let $s$ be a pixel in an image, and let $r, t, u, w$ be its four nearest neighbors, where $r, t$ are its neighbors in the Peano scan. In the extended model, the observations on remaining neighbors $u, w$ are also taken into account in the Peano scan: the image value observed on $s$ is completed with the two observations on $u$ and $w$. Initial experiments on synthetic images showed that using HMCs with such a "contextual" Peano scan allowed us to reduce the classification error by up to 17%. Our first contribution is to provide a simulation study comparing this recent HMC-CPS method to the classic PS-HMC, on one hand, and to the classic HMF-based one, on the other hand. We consider several real images as well. Our second contribution is related to the use, in the context of the CPS, of extensions of HMCs called "hidden evidential Markov chains" (HEMCs) [28]. Different "evidential" Markov chains, calling on the Dempster–Shafer theory of evidence [29], [30], have been proposed in [31], [32], [33], [34]. Here, we introduce a new model associating the CPS with HEMCs described in [28], which are particular triplet Markov chains [35], [36]. Experiments indicate the existence of situations in which the new model competes with the classic HMF, while being much faster. All experiments are unsupervised, with parameters estimated using a stochastic version of the expectation–maximization (EM) method [37], [38]. In the case of HMFs, we use the Gibbsian EM [39].

The organization of this article is as follows. In the next section, we present the contextual Peano scan and related HMC-CPS with Bayesian maximum posterior mode (MPM) segmentation and parameter estimation using SEM. We recall the classic HEMC model and extend it with a contextual scan in Section 3. Section 4 describes the experiments, and the last section contains the conclusions and perspectives.

## 2. Contextual Peano Scan and Hidden Markov chains

*2.1. Contextual Peano scan*

Let $S$ be the set of pixels of a digital image $y = (y_s)_{s \in S}$ of size $Card(S) = 2^k \times 2^k$. In the statistical segmentation framework we adopt in this study, $y = (y_s)_{s \in S}$ is a realization of a random field $Y = (Y_s)_{s \in S}$. Segmenting $y = (y_s)_{s \in S}$ consists of searching $x = (x_s)_{s \in S}$, with each $x_s$ in a finite set of classes $\Omega = \{1, ..., K\}$. Then, $x = (x_s)_{s \in S}$ is considered as a realization of a random field $X = (X_s)_{s \in S}$, and the segmentation problem is seen as a Bayesian problem of estimation of $(x_s)_{s \in S}$ from $(y_s)_{s \in S}$. For a given distribution $p(x, y)$ of $(X, Y)$, the problem of segmentation can be dealt with using a Bayesian classification, once the distribution $p(x, y)$ allows for related computation. The very elegant hidden Markov fields provide $p(x, y)$, for which Bayesian solutions are computable; however, the computational time needed can be prohibitive for numerous applications. Hidden Markov chains present alternative fast methods; however, their use requires converting the bi-dimensional set of pixels into a mono-dimensional sequence. Here, we consider the conversion based on the Peano scan presented in Figure 1.

Let us recall the use of the contextual scan introduced in [27]. Let $(1, ..., N)$ be the sequence of pixels obtained from $S$ with the Peano scan depicted in Figure 1. Let $X^N = (X_1, ..., X_N)$ be the related stochastic sequence of classes. The classic way of using the scan consists of setting $Y^N = (Y_1, ..., Y_N)$ and considering that $p(x^N, y^N)$ is a hidden Markov chain distribution. Then, both classical Maximum Posterior Mode (MPM) segmentation and Maximum a Posteriori (MAP) segmentation can be computed in a fast unsupervised way (e.g., MAP can be computed with the well-known Viterbi algorithm). We consider that the contextual scan consists of the following: For each $n = 2, ..., N - 1$, let us set $s$ as the corresponding pixel in $S$. The previous point $n - 1$ will be called $r$, and the next point $n + 1$ will be called $t$. Let us temporarily assume that $s$ is not on the border, such that it has four neighbors in $S$. Then, we associate to each $n = 2, ..., N - 1$ two pixels, $v_n$ and $w_n$, which are two neighbors of $n$ different from $r$ and $t$. Thus, each $n = s$ has four neighbors in $S$: two, $r = n - 1$ and $t = n + 1$, which belong to the Peano scan; and another two, $v_n$ and $w_n$, which do not. Then, for each $X_n$, we associate the triplet $Y_n^* = (Y_{v_n}, Y_n, Y_{w_n})$. When $s$ is on the border minus corners, there is only one neighbor, $v_n$, not lying on the Peano scan, such that $Y_n^* = (Y_{v_n}, Y_n)$. When $s$ is one of the four corners, there are two possibilities. If the scan begins or ends on the point, there is one neighbor in the scan, and one neighbor (e.g., $v_n$) outside it. Thus, $Y_n^* = (Y_{v_n}, Y_n)$. This is the case for pixels 1 and 16 in Example 2.1 below. If the scan neither begins nor ends on the point, there are no points that are neighbors in $S$ without being neighbors in the scan, so that $Y_n^* = (Y_n)$. This is the case for pixels 6 and 11 in Example 2.1 below.

Setting $Y^{*N} = (Y_1^*, \ldots, Y_N^*)$, we consider $(X^N, Y^{*N})$. As we will see, $p(x^N | y^{*N})$ is then Markovian, which allows for the implementation of Bayesian segmentation methods.

**Example 2.1.** As an example, let us consider image (b) in Figure. 1, with the Peano scan beginning in the upper left corner. Points $n = 1, \ldots, 16$ are specified in Figure. 3, and added observations are specified in Figure. 4.

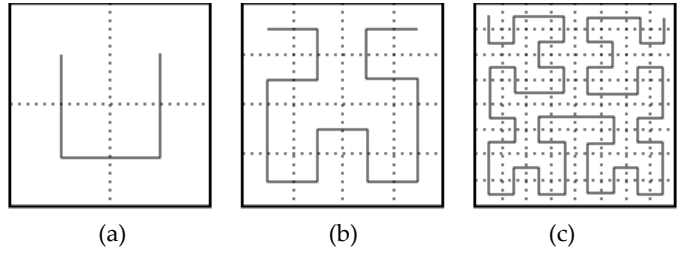

**Figure. 1.** Construction of Peano scan.

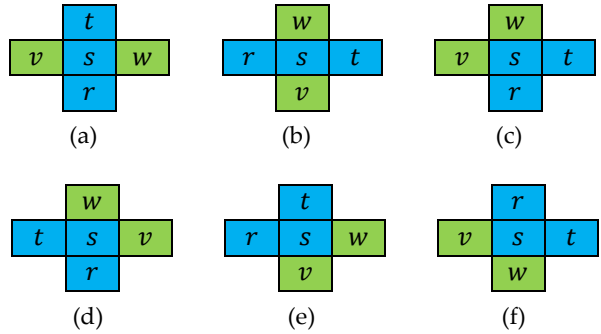

**Figure 2.** Six spatial configurations of added neighbors. In green: neighbors of $s$ are neighbors in the Peano scan; in blue: neighbors of $s$ in the set of pixels whose observations are associated with observation $y_s$ on $s$, and which are not neighbors of $s$ in the Peano scan.

| 1 | 2 | 15 | 16 |
|---|---|----|----|
| 4 | 3 | 14 | 13 |
| 5 | 8 | 9  | 12 |
| 6 | 7 | 10 | 11 |

**Figure 3.** Pixel numbers for image (b) in Figure 1, with the Peano scan beginning in the upper-left corner.

| 1 | 2 | 3 | 4 | 5 | 6 | 7 | 8 | 9 | 10 | 11 | 12 | 13 | 14 | 15 | 16 |
|---|---|---|---|---|---|---|---|---|----|----|----|----|----|----|----|
| $y_1$ | $y_2$ | $y_3$ | $y_4$ | $y_5$ | $y_6$ | $y_7$ | $y_8$ | $y_9$ | $y_{10}$ | $y_{11}$ | $y_{12}$ | $y_{13}$ | $y_{14}$ | $y_{15}$ | $y_{16}$ |
| - | - | $y_{14}$ | - | - | - | - | $y_3$ | $y_{12}$ | - | - | - | - | $y_9$ | $y_2$ | $y_{13}$ |
| $y_4$ | $y_{15}$ | $y_8$ | $y_1$ | $y_8$ | - | $y_{10}$ | $y$ | $y_{14}$ | $y_7$ | - | $y_9$ | $y_{16}$ | $y_3$ | - | - |

**Figure 4.** Observations associated with pixels 1, 2, 3, ...., 16 in Figure. 3, which are $(y_1, y_4)$, $(y_2, y_3)$, $(y_{14}, y_3, y_8)$, $(y_1, y_4)$, $(y_5, y_8)$, $y_6$, $(y_7, y_{10})$, $(y_8, y_3, y_5)$, ..., $(y_{16}, y_{13})$

Let $s$, $t$ be neighbors in $S$. If they are horizontal neighbors, let us set

$$p^h(y_t|x_s) = \sum_{x_t} p^h(x_t|x_s)p(y_t|x_t) \qquad (2.1)$$

Similarly, if they are vertical neighbors, we set

$$p^v(y_t|x_s) = \sum_{x_t} p^v(x_t|x_s)p(y_t|x_t) \qquad (2.2)$$

Thus, for $s$ in $S$ and $v$, $w$—which are neighbors of $s$ in the set of pixels but not neighbors in the Peano scan—we have

$$p(y_v, y_s, y_w|x_s) = p(y_s|x_s)p^{a(v,s)}(y_v|x_s)p^{a(w,s)}(y_w|x_s) \qquad (2.3)$$

where $a(v,s) = h$ if $v$, $s$ are horizontal neighbors, and $a(v,s) = v$ if $v$, $s$ are vertical neighbors, and the same applies for $a(w,s)$. For example, considering $(y_{14}, y_3, y_8)$ in Figure 4, we see that $3, 14$ are horizontal neighbors, while $3, 8$ are vertical neighbors. Then, we have

$$p(y_{14}, y_3, y_8|x_3) = p(y_3|x_3)p^h(y_{14}|x_3)p^v(y_8|x_3). \qquad (2.4)$$

Finally, for a given Peano scan, we associate with each pixel $s$ the two neighbors $v(s)$, $w(s)$ in the set of pixels, which are not its neighbors in the Peano scan. Numbering the Peano scan points as $(1, 2, \ldots, N)$, the related contextual Peano scan is the sequence of triplets (they are couples for points on borders, and singletons on two corners where the scan begins and ends):

$$([1, v(1), w(1)], [2, v(2), w(2)], \ldots, [N, v(N), w(N)]) \qquad (2.5)$$

The classic hidden Markov chain associated with the Peano scan has the following distribution:

$$p(x^N, y^N) = p(x_1)p(x_2|x_1) \ldots p(x_N|x_{N-1}) \, p(y_1|x_1) \ldots p(y_N|x_N) \qquad (2.6)$$

The new model we propose, called "hidden Markov chain for contextual Peano scan" (HMC-CPS), is defined as follows.
Let us consider

$$q(x^N, y^N) = p(x_1) \prod_{n=1}^{N-1} p(x_{n+1}|x_n) \prod_{n=1}^{N} p(y_n, y_{v(n)}, y_{w(n)}|x_n) \qquad (2.7)$$

It is to be noted that $q(x^N, y^N)$ is not the probability density of the pair $(X^N, Y^N)$. Nonetheless, considering the (unknown) normalizing constant

$$\kappa = \frac{1}{\sum_{x^N} \int q(x^N, y^N) dy^N}$$

we can consider the distribution

$$p(x^N, y^N) = \kappa q(x^N, y^N) \qquad (2.8)$$

**Definition 2.1** Let $S$ be a square set of pixels of dimensions $N = 2^k \times 2^k$. Let $(1, 2, \ldots, N)$ be a Peano scan (PS) of $S$, and let $([1, v(1), w(1)], [2, v(2), w(2)], \ldots, [N, v(N), w(N)])$ be the four-nearest neighbors contextual PS (4NN-CPS) associated with PS. Then, the conditional probability distribution $p(x^N|y^N)$ given from the distribution $p(x^N, y^N) = \kappa q(x^N, y^N)$ defined with (2.3), (2.7) and (2.8) will be called the "hidden Markov chain for the contextual Peano scan" (HMC-CPS) distribution.

We note that, in HMC-CPS $(X^N, Y^N)$, the chain $X^N$ is Markovian and it is also Markovian conditionally on $Y^N$, but $(X^N, Y^N)$ is not Markovian itself. As discussed in the next paragraph, $p(x_1|y^N)$ and transitions $p(x_{n+1}|x_n, y^N)$ are computable, which allow for Bayesian restorations. We note that $p(y^N|x^N)$, which is complicated to write, is not needed.

**Remark 2.1** It is possible to extend the 4NN-CPS with a richer neighborhood. Considering eight nearest neighbors in the set of pixels would lead to a contextual PS with six additional observations, except for boarding pixels, on each point in the Peano scan.

## 2.2 Bayesian MPM segmentation with HMC-CPS

Let $1, \ldots, N$ be points of a Peano scan, and let $(X^N, Y^N)$ be a HMC-CPS (2.7)-(2.8). We consider the Bayesian Marginal Posterior Mode (MPM) to be defined by

$$[\hat{s}_{MPM}(y^N) = \hat{x}^N] \Leftrightarrow \left[ p(\hat{x}_n | y^N) = \max_{x_n \in \Omega} p(x_n | y^N) \right] \quad (2.9)$$

Thus, the problem lies in computing $p(x_n | y^N)$ for $n = 1, \ldots, N$. Let us recall the following general result.

**Lemma 2.1**

Let $Z^N = (Z_1, \ldots, Z_N)$ be a stochastic chain taking its values in a finite set $\Omega$. Then,
(i) $Z^N$ is Markovian if and only if there exist $N - 1$ functions $\varphi_1, \ldots, \varphi_{N-1}$ from $\Omega^2$ to $R^+$ such that

$$p(z^N) \propto \prod_{n=1}^{N-1} \varphi_n(z_n, z_{n+1}) \quad (2.10)$$

where $\propto$ means "proportional to";

- (ii) for the verification of (2.10), $p(z_1)$ and transitions $p(z_{n+1}|z_n)$ are given from functions $\varphi_1, \ldots, \varphi_{N-1}$ with

$$p(z_1) = \frac{\beta_1(z_1)}{\sum_{z_1} \beta_1(z_1)};$$

$$\text{for } 1 < n < N, p(z_{n+1}|z_n) = \frac{\varphi_n(z_n, z_{n+1})\beta_{n+1}(z_{n+1})}{\beta_n(z_n)}, \quad (2.11)$$

where $\beta_1(z_2), \ldots, \beta_N(z_N)$ can be computed with the following backward recursion:

$$\beta_N(z_N) = 1, \text{for } n = N - 1, \ldots, 1$$

$$\beta_n(z_n) = \sum_{z_{n+1}} \varphi_n(z_n, z_{n+1})\beta_{n+1}(z_{n+1}) \quad (2.12)$$

- Once $p(z_1)$ and $p(z_{n+1}|z_n)$ are given, each $p(z_n)$ is computed with forward recursion:

$$for\ n = 2, \ldots, N, p(z_n) = \sum_{z_n} p(z_n|z_{n-1})p(z_{n-1}) \quad (2.13)$$

**Proof of Lemma 2.1**

- Let $Z^N$ be Markovian: $p(z_1, \ldots, z_N) = p(z_1)p(z_2|z_1)p(z_3|z_2) \ldots p(z_N|z_{N-1})$. Then (2.10) is satisfied by $\varphi_1(z_1, z_2) = p(z_1)p(z_2|z_1)$, $\varphi_2(z_2, z_3) = p(z_3|z_2), \ldots, \varphi_{N-1}(z_{N-1}, z_N) = p(z_N|z_{N-1})$.

- Conversely, let $p(z_1, \ldots, z_N)$ satisfy (2.10). Thus, $p(z_1, \ldots, z_N) = K\varphi_1(z_1, z_2) \ldots \varphi_{N-1}(z_{N-1}, z_N)$ with the $K$ constant, which implies that for each $n = 1, \ldots, N - 1$, we have

$$p(z_{n+1}|z_1, \ldots, z_n) = \frac{p(z_1, \ldots, z_n, z_{n+1})}{p(z_1, \ldots, z_n)} = \frac{\sum_{(z_{n+2}, \ldots, z_N)} \varphi_1(z_1, z_2) \ldots \varphi_{N-1}(z_{N-1}, z_N)}{\sum_{(z_{n+1}, \ldots, z_N)} \varphi_1(z_1, z_2) \ldots \varphi_{N-1}(z_{N-1}, z_N)} \quad (2.14)$$

which shows that $p(z_1, \ldots, z_N)$ is Markovian.
Moreover, let us set

$$\text{for } n = 1, \ldots, N - 1, \beta_n(z_n) = \sum_{(z_{n+1}, z_{n+2}, \ldots, z_N)} \varphi_n(z_n, z_{n+1}) \ldots \varphi_{N-1}(z_{N-1}, z_N)$$

On one hand, we see that

$$\beta_n(z_n) = \sum_{z_{n+1}} \varphi_n(z_n, z_{n+1})\beta_{n+1}(z_{n+1})$$

On the other hand, according to (2.14), we have

$$p(z_{n+1}|z_n) = \frac{\varphi_n(z_n, z_{n+1})\beta_{n+1}(z_{n+1})}{\beta_n(z_n)}.$$

As

$$p(z_1) = \frac{\beta_1(z_1)}{\sum_{z_1} \beta_1(z_1)}$$

(2.11) and (2.12) are verified, which ends the proof. ∎

To summarize, once functions $\varphi_1, \ldots, \varphi_{N-1}$ satisfying (2.10) are given, the distributions $p(z_1)$, $p(z_{n+1}|z_n)$ and $p(z_n)$ of the related Markov chain $Z^N = (Z_1, \ldots, Z_N)$ are computable. Thus, it is sufficient to define $\varphi_1, \ldots, \varphi_{N-1}$ satisfying (2.10), which will highly simplify different model introductions in practice. Let us return to the conditional Markov chain for contextual Peano scan distribution $p(x^N|y^N)$, specified in Definition 2.1. We can say that $p(x^N|y^N) \propto q(x^N, y^N)$, and thus,

$$p(x^N|y^N) \propto \prod_{n=1}^{N-1} \varphi_n(x_n, x_{n+1}, y^N) \tag{2.15}$$

with

$$\varphi_1(x_1, x_2, y^N) = p(x_1, x_2)p(y_1, y_{v(1)}, y_{w(1)}|x_1)p(y_2, y_{v(2)}, y_{w(2)}|x_2)$$
$$\varphi_2(x_2, x_3, y^N) = p(x_3|x_2)p(y_3, y_{v(3)}, y_{w(3)}|x_3)$$
$$\ldots \tag{2.16}$$
$$\varphi_{N-1}(x_{N-1}, x_N, y^N) = p(x_N|x_{N-1})p(y_N, y_{v(N)}, y_{w(N)}|x_N)$$

Finally, functions $\varphi_1, \ldots, \varphi_{N-1}$ satisfying (2.10) are of the form

$$\varphi_1(x_1, x_2, y^N) = \varphi_1(x_1, x_2, y_1, y_{v(1)}, y_{w(1)}, y_2, y_{v(2)}, y_{w(2)})$$
$$\varphi_2(x_2, x_3, y^N) = \varphi_2(x_2, x_3, y_3, y_{v(3)}, y_{w(3)})$$
$$\ldots \tag{2.17}$$
$$\varphi_{N-1}(x_{N-1}, x_N, y^N) = \varphi_{N-1}(x_{N-1}, x_N, y_N, y_{v(N)}, y_{w(N)})$$

and thus, they are easy to compute. Then, $p(x_1|y^N)$ and transitions $p(x_{n+1}|x_n, y^N)$ are computable following Lemma 2.1, which gives marginal distributions $p(x_n|y^N)$ of $p(x^N|y^N)$ and allows the use of MPM (2.9).

**Remark 2.2** The pair $(X^N, Y^N)$ has a complex and only partially known structure. In particular, neither $p(x^N)$ nor $p(y^N|x^N)$ is Markovian in general. In addition, for $p(y_n|x_n)$ Gaussian, $p(y^N|x^N)$ is not Gaussian in general. However, this is of little importance because what is important is that $p(x^N|y^N)$ is Markovian and known up to a constant, which makes $p(x_n|y^N)$ computable.

**Example 2.2** Let us specify $\varphi_1, \varphi_2, \ldots, \varphi_{15}$ used in Example 2.1. According to Fig. 3 and (2.3), (2.15), we have

$$\varphi_1(x_1, x_2, y^N) = p(x_1, x_2)p(y_1|x_1)p^v(y_4|x_1)p(y_2|x_2)p^h(y_{15}|x_2)$$
$$\varphi_2(x_2, x_3, y^N) = p(x_3|x_2)p(y_3|x_3)p^v(y_8|x_3)p^h(y_{14}|x_3)$$
$$\varphi_3(x_3, x_4, y^N) = p(x_4|x_3)p(y_4|x_4)p^v(y_1|x_4)$$
$$\ldots \tag{2.18}$$
$$\varphi_{13}(x_{13}, x_{14}, y^N) = p(x_{14}|x_{13})p(y_{14}|x_{14})p^v(y_9|x_{14})p^h(y_3|x_{14})$$
$$\varphi_{14}(x_{14}, x_{15}, y^N) = p(x_{15}|x_{14})p(y_{15}|x_{15})p^h(y_2|x_{15})$$
$$\varphi_{15}(x_{15}, x_{16}, y^N) = p(x_{16}|x_{15})p(y_{16}|x_{16})p^v(y_{13}|x_{16})$$

*2.3. Parameter estimation*

Let us suppose that $p(x^N, y^N)$ is a classic hidden Markov chain (CHMC) distribution, with Gaussian $p(y^N|x^N)$ and two different transitions depending on whether it applies to horizontal neighbors or vertical neighbors in the original image. For $K$ classes $\Omega = \{1, ..., K\}$, the parameters are as follows: $K^2$ probabilities $p^h = (p_{ij}^h)_{1 \leq i,j \leq K}$, with $p_{ij}^h = p(x_{t+1} = j, x_t = i)$ for $t, t+1$ neighbors in the chain and horizontal neighbors in the image, $K^2$ probabilities $p^v = (p_{ij}^v)_{1 \leq i,j \leq K}$, with $p_{ij}^v = p(x_{t+1} = j, x_t = i)$ for $t, t+1$ neighbors in the chain and vertical neighbors in the image, and $K$ means $m = (m_i)_{1 \leq i \leq K}$ and $K$ variances $\sigma^2 = (\sigma_i^2)_{1 \leq i \leq K}$ of the $K$ Gaussian distributions ($p(y_s|x_s = i))_{1 \leq i \leq K}$. Let us consider the HMC-CPS: by choosing $p^h(x_t|x_s)$ and $p^v(x_t|x_s)$ intervening in $p(y_u, y_s, y_w|x_s)$ to be those related to $p_{ij}^h$ and $p_{ij}^v$ respectively, the new proposed model uses exactly the same parameters as the CHMC. Thus, the problem is to estimate $\theta = (p^h, p^v, m, \sigma^2)$ from the observed image $Y^N = y^N$ alone. However, contrary to the CHMC model, the EM algorithm cannot be used for estimating the parameters. Indeed, according to (2.7) and (2.8), the joint likelihood $p(x^N, y^N, \theta)$ is only computable up to a constant $\kappa$, which depends on $\theta$. As such, it is not possible in general to compute $\underset{\theta}{\arg\max} E[\log(p(x^N, y^N, \theta))|y^N, \theta^q]$ for each iteration $q$.

Nonetheless, as $p(x^N|y^N, \theta^q)$ is computable and, being Markovian, it is possible to sample from it, we can use stochastic EM (SEM), which is performed as follows.

- Initialize the parameters $\theta^0 = (p^{h,0}, p^{v,0}, m^0, \sigma^{2,0})$ with some simple method;
- Compute $\theta^{q+1} = (p^{h,q+1}, p^{v,q+1}, m^{q+1}, \sigma^{2,q+1})$ from current $\theta^q = (p^{h,q}, p^{v,q}, m^q, \sigma^{2,q})$ and $y^N$;
- Sample $x^{N,q+1} = (x_1^{q+1}, ..., x_N^{q+1})$ according to the Markov distribution $p(x^N|y^N, \theta^q)$;
- Let $H^{q+1}$ be the set of couples $(n, n+1)$, with $n$ and $n+1$ horizontal neighbors in the set of pixels, and let $V^{q+1}$ be the set of couples $(n, n+1)$, with $n$ and $n+1$ vertical neighbors in the set of pixels. Let $S^{i,q+1}$ be the set of points $n$ such that $x_n^{q+1} = i$. Set

$$p_{ij}^{h,q+1} = \frac{\sum_{(n,n+1) \in H^{q+1}} 1_{[x_n^{q+1}=i, x_{n+1}^{q+1}=j]}}{|H^{q+1}|}, \text{ similar for } p_{ij}^{v,q+1} \quad (2.19)$$

$$m_i^{q+1} = \frac{\sum_{n \in S^{i,q+1}} y_n}{|S^{i,q+1}|}, \sigma_{i,q+1}^2 = \frac{\sum_{n \in S^{i,q+1}} (y_n - m_i^{q+1})^2}{|S^{i,q+1}|} \quad (2.20)$$

A stopping criterion is adapted to each case considered. Indeed, the Markov chain ($\theta^q$) obtained in this way is very complex and its stabilization depends on different factors, particularly on the noise level. To the best of our knowledge, as in the EM case, there are no general theoretical results specifying its asymptotic behavior.

**Remark 2.3** One can use a simpler approximated SEM by treating $p(x^N, y^N)$ as a CHMC and sampling from its posterior law $p^{CHMC}(x^N|y^N, \theta^q)$ instead of the real $p(x^N|y^N, \theta^q)$; we programmed it and it produces slightly worse results.

**3. Contextual Peano Scan and Hidden Evidential Markov chains**

*3.1. MPM restoration with hidden Triplet Markov chains*

Let us briefly examine particular triplet Markov chains, named hidden Triplet Markov chains. Let $X^N = (X_1, ..., X_N)$ be the stochastic sequence of classes, $Y^N = (Y_1, ..., Y_N)$ the sequence of observations on the pixels and $U^N = (U_1, ..., U_N)$ a third stochastic sequence, with each $U_n$ taking its values in $\Lambda = \{1, ..., L\}$. Let $T^N = (T_1, ..., T_N)$, with $T_n = (X_n, U_n, Y_n)$, be a triplet Markov chain. Then, $X^N$ can be estimated from $Y^N$ like in the classic HMC case. As $U^N$ is arbitrary, the family of TMCs is a very general one. As $p(t^N)$ is written $p(t^N) = p(t_1)p(t_2|t_1) ... p(t_N|t_{N-1})$, it has the form

$$p(t^N) = \varphi_1(x_1, x_2, u_1, u_2, y_1, y_2) ... \varphi_{N-1}(x_{N-1}, x_N, u_{N-1}, u_N, y_{N-1}, y_N) \quad (3.1)$$

with

$$\varphi_1(x_1, x_2, u_1, u_2, y_1, y_2) = p(t_1, t_2)$$
$$\varphi_2(x_2, x_3, u_2, u_3, y_2, y_3) = p(t_3|t_2)$$
$$... \quad (3.2)$$

$$\varphi_{N-1}(x_{N-1}, x_N, u_{N-1}, u_N, y_{N-1}, y_N) = p(t_N|t_{N-1})$$

Then, applying Lemma 2.1 to $Z^N = (X^N, U^N)$ and $p(z^N|y^N)$, we can compute $p(z_n|y^N) = p(x_n, u_n|y^N)$ for each $n = 1, \ldots, N$, which finally gives

$$p(x_n|y^N) = \sum_{u_n \in \Omega} p(x_n, u_n|y^N) \tag{3.3}$$

Allowing the application of Bayesian MPM segmentation (2.9). Throughout this article, we will consider a particular TMC, satisfying

$$p(t_1) = p(x_1, u_1)p(y_1|x_1)$$
$$p(t_{n+1}|t_n) = p(x_{n+1}, u_{n+1}|x_n, u_n)p(y_{n+1}|x_{n+1}) \tag{3.4}$$

**Remark 3.1** We may observe that (3.4) is a simplified TMC, obtained by assuming $p(x_{n+1}, u_{n+1}|x_n, u_n, y_n) = p(x_{n+1}, u_{n+1}|x_n, u_n)$, and $p(y_{n+1}|x_n, u_n, y_n, x_{n+1}, u_{n+1}) = p(y_{n+1}|x_{n+1})$. Setting $W_n = (X_n, U_n)$, we can see that $T^N = (W^N, Y^N)$ has the structure of a very classical HMC: $W^N$ is Markovian and $p(y_n|w^N) = p(y_n|w_n)$. In addition, it satisfies $p(y_n|w_n) = p(y_n|x_n, u_n) = p(y_n|x_n)$. This shows that such TMCs are not much different from classical HMCs, so that different computer programs related to HMCs can be adapted to these TMCs with slight modifications.

**Remark 3.2** Let us note the great generality of triplet Markov chains, even in their particular form (3.4). In particular, $(X^N, U^N)$ is Markov, but $X^N$ is not necessarily Markov. Moreover, in TMCs, $U^N$ is free, and the only condition is that each $U_n$ takes its values in a finite not-too-large set. In particular, it can be multivariate: $U_n = (U_n^1, \ldots, U_n^I)$, with each $U_n^i$ modeling a particular aspect of the problem dealt with.

*3.2. Models combining contextual Peano scan with triplet Markov chains*

Let $S$ be a square set of pixels of dimensions $N = 2^k \times 2^k$. Let $(1, 2, \ldots, N)$ be a Peano scan (PS) of $S$, and let $([1, v(1), w(1)], [2, v(2), w(2)], \ldots, [N, v(N), w(N)])$ be the four nearest neighbors' contextual PS (4NN-CPS) associated with PS. Let us consider a TMC of the form (3.1). We limit our presentation to particular TMCs $(X^N, U^N, Y^N)$ satisfying (3.4). Setting $Z_n = (X_n, U_n)$ for $n = 1, \ldots, N$, $T^N = (Z^N, Y^N)$ is a classic HMC, and thus, in a similar manner to Section 2.1, we can extend this model to contextual Peano scan. We define the "conditional triplet Markov chain for contextual Peano scan" (CTMC-CPS) distribution by setting

$$p^h(y_t|z_s) = \sum_{x_t, u_t} p^h(x_t, u_t|z_s)p(y_t|x_t) \tag{3.5}$$

if $s, t$ are horizontal neighbors. Furthermore,

$$p^v(y_t|z_s) = \sum_{x_t, u_t} p^v(x_t, u_t|z_s)p(y_t|x_t) \tag{3.6}$$

if they are vertical neighbors. Then, using Definition 2.1, we obtain

$$p(y_v, y_s, y_w|z_s) = p(y_s|z_s)p^{a(v,s)}(y_v|z_s)p^{a(w,s)}(y_w|z_s) \tag{3.7}$$

with $a(v, s) = h$ if $v, s$ are horizontal neighbors, and $a(v, s) = v$ if $v, s$ are vertical neighbors, and finally, we obtain the CTMC-CPS distribution:

$$p(z^N, y^N) = \kappa p(z_1) \prod_{n=1}^{N-1} p(z_{n+1}|z_n) \prod_{n=1}^{N} p(y_n, y_{v(n)}, y_{w(n)}|z_n) = \kappa q(z^N, y^N) \tag{3.8}$$

with $v(n), w(n)$ being the two neighbors of pixel $n$ in the set of pixels, which are not its neighbors in the Peano scan.

Similarly to Section 2.1, $p(z^N, y^N)$ is not Markovian here; thus, $\kappa$ is not computable in general. However, distribution $p(z^N|y^N) = p(x^N, u^N|y^N)$ being Markovian, $p(x_n, u_n|y^N)$ is computable, and thus, $p(x_n|y^N)$ is also computable by applying (3.3). Thus, the problem is similar to that in Section 2.2; the only difference is that, here, one considers $(x_n, u_n)$ instead of $x_n$ and, once $p(x_n, u_n|y^N)$ is computed, one applies (3.3) to compute $p(x_n|y^N)$ and to apply MPM. For parameter estimation, the SEM from Section 2.3 can be used, with $p(z^N, y^N)$ instead of $p(x^N, y^N)$. For $\Omega = \{1, \ldots, K\}$ and $\Lambda = \{1, \ldots, L\}$, the parameters are as follows: $K^2L^2$ probabilities $p^h = (p^h_{ijkl})_{\substack{1\leq i,j\leq K \\ 1\leq k,l\leq L}}$, with $p^h_{ijkl} = p(x_{t+1} = j, u_{t+1} = l, x_t = i, u_t = k)$ for $t, t+1$ neighbors in the chain and horizontal neighbors in the image, $K^2L^2$ probabilities $p^v = (p^v_{ijkl})_{\substack{1\leq i,j\leq K \\ 1\leq k,l\leq L}}$, with $p^v_{ijkl} = p(x_{t+1} = j, u_{t+1} = l, x_t = i, u_t = k)$ for $t, t+1$ neighbors in the chain and vertical neighbors in the image, and $K$ means $m = (m_i)_{1\leq i\leq K}$ and $K$ variances $\sigma^2 = (\sigma_i^2)_{1\leq i\leq K}$ of the $K$ Gaussian distributions ($p(y_s|x_s = i))_{1\leq i\leq K}$.

*3.3 Hidden evidential Markov chain for contextual Peano scan*

Hidden evidential Markov chains (HEMCs) are particular triplet Markov chains $(X^N, U^N, Y^N)$ satisfying (3.4), in which each $U_n$ takes its values in $2^\Omega$, the power set of $\Omega$. To define $p(t^N)$, we first define $m(u^N)$, which is a "basic belief assignment" (bba) $m$ defined on $(2^\Omega)^N$ with

$$m(u_1, \dots, u_N) = m(u_1) \prod_{i=1}^{N-1} m(u_{n+1}|u_n) \tag{3.9}$$

and null on $2^{(\Omega^N)} - (2^\Omega)^N$. This simply means that, for each $n = 1, \dots, N-1$ and $u_n \epsilon 2^\Omega$, $u_{n+1} \to m(u_{n+1}|u_n)$ is a bba on $2^\Omega$. To simplify, a bba $m$ on $2^\Omega$ in this article will be seen as a probability satisfying $m(\emptyset) = 0$.

Having (3.9), evidential Markov chain (EMC) distribution is the Markovian distribution $p(x^N, u^N)$ obtained with normalization, i.e., with its division by $\sum_{(x^N,u^N)} \varphi(x^N, u^N)$, of the function $\varphi$ defined in $\Omega^N \times (2^\Omega)^N$ with

$$\varphi(x^N, u^N) = \varphi_1(x_1, u_1, x_2, u_2) \prod_{n=2}^{N-1} \varphi_n(u_n, x_{n+1}, u_{n+1}) \tag{3.10}$$

where

$$\varphi_1(x_1, u_1, x_2, u_2) = 1_{[x_1 \in u_1, x_2 \in u_2]} m(u_1) m(u_2|u_1) \tag{3.11}$$

$$\varphi_n(u_n, x_{n+1}, u_{n+1}) = 1_{[x_{n+1} \in u_{n+1}]} m(u_{n+1}|u_n), \text{for } n = 2, \dots, N-1 \tag{3.12}$$

Setting $Card(u_n) = |u_n|$ and applying Lemma 2.1, we find, after calculation,

$$p(x_1, u_1) = \frac{|u_1| m(u_1) \sum_{(x_2, u_2)} \varphi_1(x_1, u_1, x_2, u_2) \beta_2^*(u_2)}{\sum_{u_1} |u_1| m(u_1) \sum_{(x_2, u_2)} \varphi_1(x_1, u_1, x_2, u_2) \beta_2^*(u_2)} = p(u_1) p(x_1|u_1) \tag{3.13}$$

$$p(x_{n+1}, u_{n+1}|x_n, u_n) = p(x_{n+1}, u_{n+1}|u_n) = p(u_{n+1}|u_n) p(x_{n+1}|u_{n+1})$$

with

$$p(x_n|u_n) = \frac{1_{[x_n \in u_n]}}{|u_n|}, \text{for } n = 1, \dots, N \tag{3.14}$$

$$for\ n = 1, \dots, N-1, p(u_{n+1}|u_n) = \frac{|u_{n+1}| m(u_{n+1}|u_n) \beta_{n+1}^*(u_{n+1})}{\sum_{u_{n+1}} |u_{n+1}| m(u_{n+1}|u_n) \beta_{n+1}^*(u_{n+1})} \tag{3.15}$$

where $\beta_N^*(u_N), \dots, \beta_2^*(u_2)$ are computed with backward induction:

$$\beta_N^*(u_N) = 1$$

$$for\ n = 2, \dots, N-1, \beta_n^*(u_n) = \sum_{(x_{n+1}, u_{n+1})} \varphi_n(u_n, x_{n+1}, u_{n+1}) \beta_{n+1}^*(u_{n+1}) \tag{3.16}$$

Having EMC $p(x^N, u^N)$, we define hidden EMC (HEMC) $p(x^N, u^N, y^N)$ with

$$p(x^N, u^N, y^N) = p(u_1) p(x_1|u_1) p(y_1|x_1) \prod_{n=1}^{N-1} p(u_{n+1}|u_n) \, p(x_{n+1}|u_{n+1}) \, p(y_{n+1}|x_{n+1}) \tag{3.17}$$

with

$$p(x_n|u_n) = \frac{1_{[x_n \in u_n]}}{|u_n|} \text{ for } n = 1, \dots, N \tag{3.18}$$

As HEMCs satisfy (3.4), to construct the hidden evidential Markov chain for the contextual Peano scan (HEMC-CPS), setting $Z_n = (X_n, U_n)$, we can use (3.5) and (3.6), which become

$$p^h(y_t|u_s) = \sum_{x_t, u_t} p^h(u_t|u_s) p(x_t|u_t) p(y_t|x_t) \qquad (3.19)$$

if $s, t$ are horizontal neighbors;

$$p^v(y_t|u_s) = \sum_{x_t, u_t} p^v(u_t|u_s) p(x_t|u_t) p(y_t|x_t) \qquad (3.20)$$

if they are vertical neighbors and (3.7), which becomes

$$p(y_v, y_s, y_w|z_s) = p(y_s|x_s) p^{a(v,s)}(y_v|u_s) p^{a(w,s)}(y_w|u_s) \qquad (3.21)$$

to obtain the HEMC-CPS distribution:

$$p(z^N, y^N) = \kappa p(x_1) p(u_1|x_1) \prod_{n=1}^{N-1} p(u_{n+1}|u_n) p(x_{n+1}|u_{n+1}) \prod_{n=1}^{N} p(y_n, y_{v(n)}, y_{w(n)}|z_n) \qquad (3.22)$$

with $v(n), w(n)$, the two neighbors of pixel $n$ in the set of pixels, which are not its neighbors in the Peano scan and $p(x_n|u_n)$ satisfying (3.18) for $n = 1, \ldots, N$.

In practice, and in the experiments below, one often uses the following simple bba:

$$m(u_{n+1}|u_n) = 0 \text{ for } u_{n+1} \text{ outside of } \{\{1\}, \{2\}, \ldots, \{K\}, \{1, \ldots, K\}\} \qquad (3.23)$$

Thus, such a hidden evidential Markov chain is a light extension of the classic HMC, as we find the latter again for $m(u_{n+1} = \{1, \ldots, K\}|u_n) = 0$ for each $n$ and each $u_n$. For $\Omega = \{1, \ldots, K\}$ and $m(u_{n+1}|u_n)$ satisfying (3.23), the parameters are as follows: $(K + 1)^2$ probabilities $p^h = (p_{kl}^h)_{1 \leq k, l \leq K+1}$, with $p_{kl}^h = p(u_{t+1} = l, u_t = k)$ for $t, t + 1$ neighbors in the chain and horizontal neighbors in the image, $(K + 1)^2$ probabilities $p^v = (p_{kl}^v)_{1 \leq k, l \leq K+1}$, with $p_{kl}^v = p(u_{t+1} = l, u_t = k)$ for $t, t + 1$ neighbors in the chain and vertical neighbors in the image, and $K$ means $m = (m_i)_{1 \leq i \leq K}$ and $K$ variances $\sigma^2 = (\sigma_i^2)_{1 \leq i \leq K}$ of the $K$ Gaussian distributions ($p(y_s|x_s = i))_{1 \leq i \leq K}$. Let us note that $p(x_n|u_n)$ is not a parameter, as it is not free and must satisfy (3.18). Finally, these parameters can be estimated using the SEM from Section 2.3 by replacing (2.19) with

$$p_{kl}^{h,q+1} = \frac{\sum_{(n,n+1) \in H^{q+1}} \sum_{i \in \Omega} \sum_{j \in \Omega} 1_{[u_n^{q+1}=k,\, x_n^{q+1}=i,\, x_{n+1}^{q+1}=j,\, u_{n+1}^{q+1}=l]}}{|H^{q+1}|}, \text{ similar for } p_{kl}^{v,q+1} \qquad (3.24)$$

**Remark 3.3** Let us briefly specify the motivation of introducing theory of evidence (TE) in an image segmentation context. For the case of two classes $\{1, 2\}$, we have noticed—see [28], [40] and references therein—that, in small-sized areas, the a posteriori law of the evidential variable loads $p(u_{n+1} = \{1,2\}|u_n, y^N)$ at the expense of $p(u_{n+1} = \{1\}|u_n, y^N)$ and $p(u_{n+1} = \{2\}|u_n, y^N)$, which has the effect of reducing the role of context. Then, this reduction appears to be beneficial for segmentation. Similarly, in large-sized areas, it loads $p(u_{n+1} = \{1\}|u_n, y^N)$ and $p(u_{n+1} = \{2\}|u_n, y^N)$ to the detriment of $p(u_{n+1} = \{1,2\}|u_n, y^N)$, which also improves segmentation. In cases where both types of area exist in an image, this mechanism improves (sometimes significantly) the segmentation obtained with the classic Markovian model, without the third evidential variable, which treats all areas in the same way. It is worth noting that the value of DS fusion in Markovian segmentation in a non-stationary data case has been found empirically, and we have no theoretical justification to propose at the moment. Why classical parameter estimation in evidential Markov models, which are classical triplet Markov models, provides parameters allowing for the double property mentioned above is not clear. Searching for theoretical justifications seems to be a challenging problem.

# 4. Experiments

*4.1. Segmentation of Synthetic images*

This section shows some results of hand-drawn noisy images subjected to unsupervised segmentation based on two models presented in this study, as well as on an HMC and on an HMF for reference. We estimate the parameters using the SEM algorithms from Sections 2.3 and 3.3, with initializations obtained from a k-means algorithm run on the observed data. HMF parameters are estimated with the Gibbsian EM algorithm, with its initialization also obtained from a k-means algorithm. According to the results presented in Figure 5, we can propose the following:

The main conclusion is that in the frame of classic hidden Markov chains, the contextual Peano scan-based method is systematically and significantly more efficient than the classic Peano scan one. The relative average gain is about 16%. Another important conclusion is that in some, though not all, cases, the efficiency of the CPS-based method is comparable to that of the hidden Markov field-based one. This is the case in the "Zebra", "Zebra with target" or "Spaghetti" images. In these three cases, Markov field-based segmentation appears to be worse.

The value of extending hidden Markov to evidential Markov is less striking. However, it can significantly improve the results obtained with hidden Markov chains in the case of images containing simultaneously large areas and fine details. This means that extending HMC-CPS to HEMC-CPS should improve the model in the same fashion. This is confirmed through the experiments, particularly for "Tree" and "Zebra with letter", for which HEMC-CPS does indeed improve HMC-CPS, while being inferior to HMF. The same occurs for "Digital" and "Nazca" where HEMC-CPS outperforms both HMC-CPS and HMF. It should also be noted that HEMC-CPS outperforms the hidden evidential Markov field model (HEMF), presented in [40], for the "Nazca" image (HEMF scores 6.7% of error, and HEMC-CPS 5.4%). Moreover, if we look at the "Spaghetti" and "Squares" images, we can see something interesting: When considering error rates, HMF is the best model, followed by HMC-CPS and then HEMC-CPS. However, some parts of the images (like the spaghetti shapes in "Spaghetti" and the three fine squares that are numbered 4,5,6 from the center of the image "Squares") are segmented with more details using HEMC-CPS, while HMC-CPS seems to miss some of the details and HMF misses even more. Finally, in theory, HEMC-CPS is strictly more general than HMC-CPS; however, in practice, it sometimes performs worse than HMC-CPS, as we can see in "Zebra" and "Zebra with target". Of course, parameter estimation is more difficult in HEMC-CPS than in HMC-CPS, which is a possible explanation.

*4.2. Segmentation of Real images*

We chose two real grayscale images and segment them using the same four methods: HMC-PS, HMC-CPS, HEMC-CPS and HMF. From the results presented in Figure 6, we can make several observations:

For "Fractured Bones", while HMF has a more satisfying segmentation of the bones, it is not able to segment the fractured zone, while this zone appears in HMC-CPS segmentation. Both HEMC-CPS and HMC-PS perform quite badly on this image.

For "Tree in the field", if we look at the segmentation of the fine tree branches, HMC-CPS segmentation seems the least detailed. HMC-PS seems more detailed than HMC-CPS, but less than HMF and HEMC-CPS. Moreover, it is very difficult to decide between the two latter models. In particular, both segmentations of fine branches seem almost identical. This is encouraging as it demonstrates two points: first, HEMC-CPS can be useful with respect to HMC-CPS in real situations; second, in the same kind of real situations, HEMC-CPS-based segmentation can be as efficient as HMF-based segmentation.

Finally, we will discuss the computational complexity of the different methods considered. MPM restoration in HMC and HEMC has a time complexity in $O(N)$, where $N$ is the length of the sequence, if we make the assumption that the number of hidden states is negligible compared to $N$. Under the same assumption, MPM in HMC-CPS and HEMC-CPS also has a time complexity in $O(N)$ if the neighborhood considered is not too large. For the parameter estimation with SEM, the time complexity for one iteration of the algorithm is also $O(N)$ for HMC, HEMC, HMC-CPS and HEMC-CPS. If we compare with HMF, the exact MPM time complexity is not calculable. However, one full iteration of Gibbs sampling has a time complexity of $O(N)$. Considering that for approximate MPM, we need to have several simulations, and each simulation is obtained by running several iterations of Gibbs sampling, the total number of Gibbs iterations becomes non-negligible in front of $N$ for a good MPM approximation. The same can be said for the Gibbsian EM algorithm in which for each iteration, we need several simulations, each of them requiring a number of Gibbs sampling iterations. In the experiments, this made HMF-based MPM about a hundred times longer than HMC-CPS- and HEMC-CPS-based ones.

To conclude, we only identified one case where the extension from Markov to evidential Markov is relevant, which is the segmentation of images combining fine details (about one pixel wide) and low noise level, but finding real-world applications presenting these particular properties is challenging. Nonetheless, we showed that both HMC-CPS and HEMC-CPS can be useful as they both improve upon HMC, while being a faster and still relevant alternative to HMF-based MPM, even possibly in cases of real images. Furthermore, other cases where HEMC-CPS significantly outperforms HMC-CPS may remain to be discovered.

| Gaussian noise | Image | HMC-PS | HMC-CPS | HEMC-CPS | HMF |
|---|---|---|---|---|---|
| 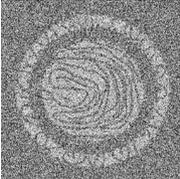 | 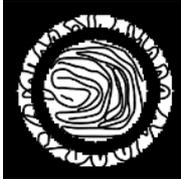 | 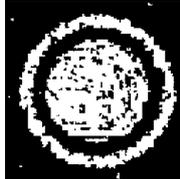 | 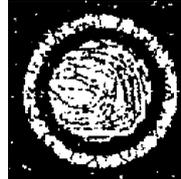 | 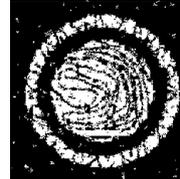 | 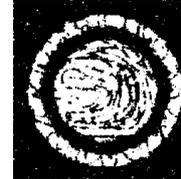 |
| $N(0,1), N(1,1)$ | Spaghetti | 0.122 | 0.091 | 0.092 | 0.089 |
| 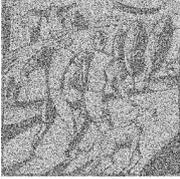 | 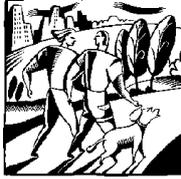 | 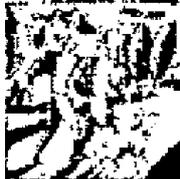 | 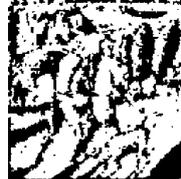 | 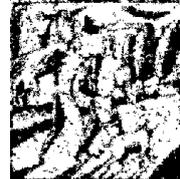 | 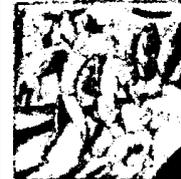 |
| $N(0,1), N(1,1)$ | Walk | 0.153 | 0.139 | 0.141 | 0.116 |
| 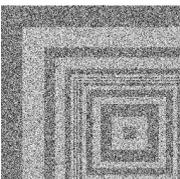 | 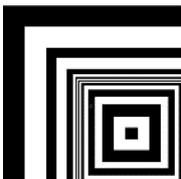 | 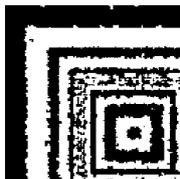 | 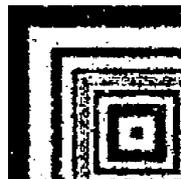 | 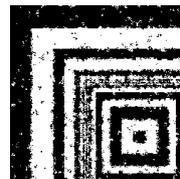 | 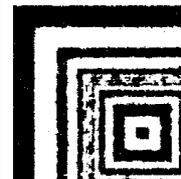 |
| $N(0,1), N(1,1)$ | Squares | 0.095 | 0.074 | 0.089 | 0.061 |
| 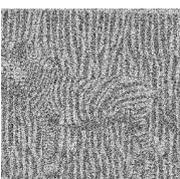 | 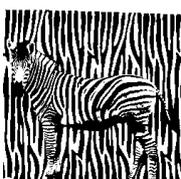 | 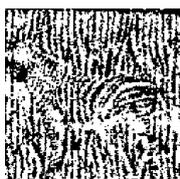 | 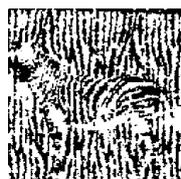 | 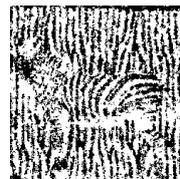 | 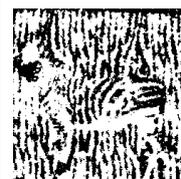 |
| $N(0,1), N(1,1)$ | Zebra | 0.201 | 0.178 | 0.187 | 0.180 |
| 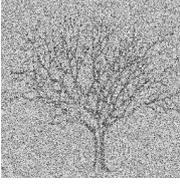 | 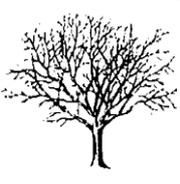 | 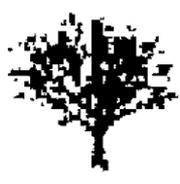 | 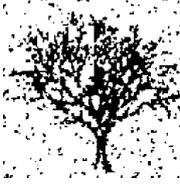 | 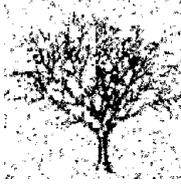 | 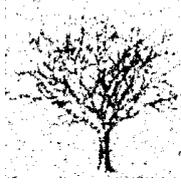 |
| $N(0,1)$, $N(1,1)$ | Tree | 0.172 | 0.149 | 0.115 | 0.088 |
| 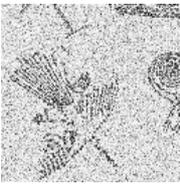 | 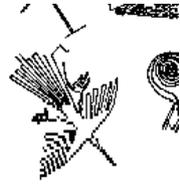 | 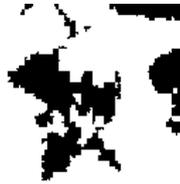 | 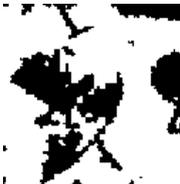 | 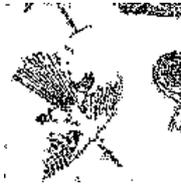 | 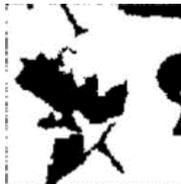 |
| $N(0,1)$, $N(2,1)$ | Nazca | 0.154 | 0.158 | 0.054 | 0.128 |
| 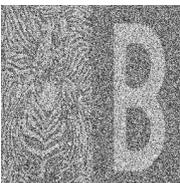 | 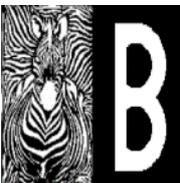 | 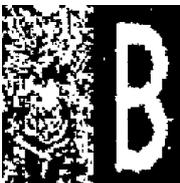 | 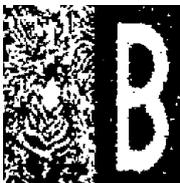 | 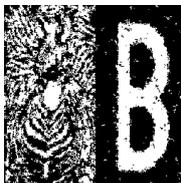 | 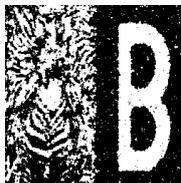 |
| $N(0,1)$, $N(1,1)$ | Zebra with letter | 0.161 | 0.141 | 0.131 | 0.120 |
| 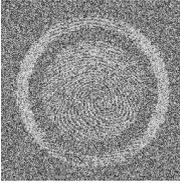 | 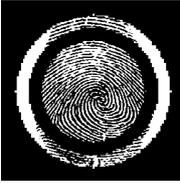 | 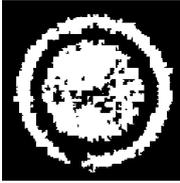 | 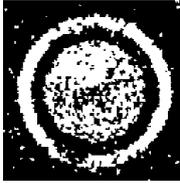 | 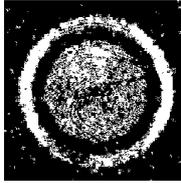 | 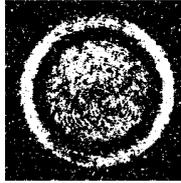 |
| $N(0,1)$, $N(1,1)$ | Digital | 0.170 | 0.155 | 0.120 | 0.125 |
| 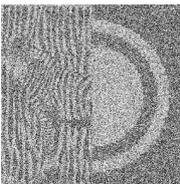 | 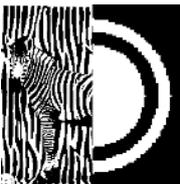 | 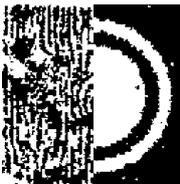 | 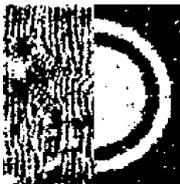 | 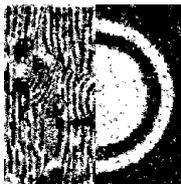 | 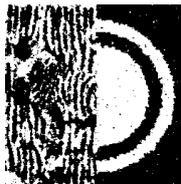 |
| $N(0,1)$, $N(1,1)$ | Zebra with target | 0.138 | 0.117 | 0.131 | 0.121 |

**Figure 5.** Unsupervised segmentation of seven images with Gaussian noise using four Bayesian MPM methods: classic hidden Markov chains and classic Peano scan (HMC-PS), classic HMC and contextual Peano scan (HMC-CPS), evidential HMC and contextual Peano scan (HEMC-CPS), and classic hidden Markov field (HMF).

| Image | | | | |
|---|---|---|---|---|
| 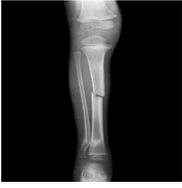 | 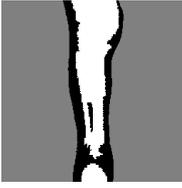 | 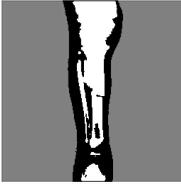 | 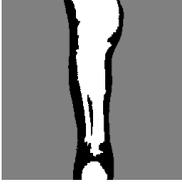 | 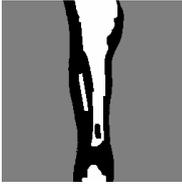 |
| Fractured bones | HMC-PS | HMC-CPS | HEMC-CPS | HMF |
| 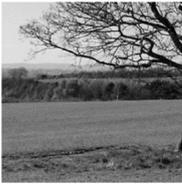 | 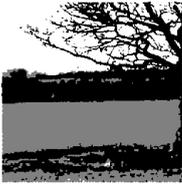 | 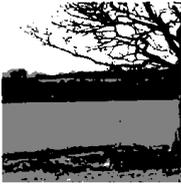 | 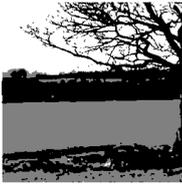 | 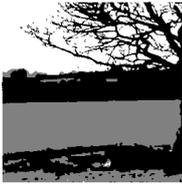 |
| Tree in the field | HMC-PS | HMC-CPS | HEMC-CPS | HMF |

**Figure 6.** Unsupervised segmentation of two real grayscale images using four Bayesian MPM methods: classic hidden Markov chains and classic Peano scan (HMC-PS), classic HMC and contextual Peano scan (HMC-CPS), evidential HMC and contextual Peano scan (HEMC-CPS), and classic hidden Markov field (HMF).

## 5. Conclusions and Perspectives

Using hidden Markov chains to design fast unsupervised image segmentation methods via the Peano scan (PS) is feasible. In this article, we investigated how the introduction of the "contextual" Peano scan (CPS) allows for the improvement of classical methods. According to the detailed experiments, introducing the CPS can significantly improve Bayesian unsupervised segmentations, with the average improvement measured in the error ratio being about 15%. Furthermore, in some situations, the quality of CPS-based segmentations can approach that of classic hidden Markov field-based ones, while being much faster. Moreover, hidden Markov chains are generalizable to pairwise and triplet Markov ones, and classic PS-based methods are generalizable to more sophisticated models; thus, it is the same for CPS-based models. We tested one of these models based on the theory of evidence. General conclusions are more difficult to draw; however, its value becomes apparent in particular situations, such as when both large- and small-sized areas are present in the hidden class image. The efficiency of MPM based on such evidential models can even match that of MPM based on HMFs in the case of real images, while being much faster.

In this study, we considered non-causal inferences. For $1 < n < N$, we obtained $X_n$, and $Y_n^* = (Y_n, Y_n')$, with $Y_n' = (Y_{v_n}, Y_{w_n})$, where $v_n, w_n$ are pixel neighbors of $n$, albeit outside of the scan. Stochastic inferences in such models are illustrated by the dependency graph shown in Fig. 7 (a). However, similar models remain valid for causal inference. The corresponding graph is the directed graph shown in Fig. 7 (b). Thus, the general structure of the family of models considered is suitable for both causal and non-causal situations.

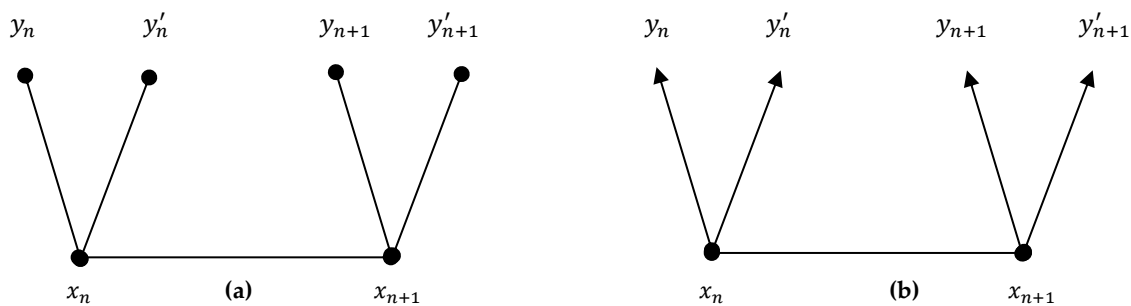

**Figure 7.** Undirected dependence graph (a) and directed one (b), related to non-causal and causal situations, respectively.

In terms of research perspectives, let us note that the flexibility of the Peano scan allows for its use in 3D images [41] and sequences of images [18], which are significantly more complex than the fixed images considered in this study, and in which using HMF techniques is very time-consuming. Thus, the proposed CPS-based techniques are likely to be of interest in such complex situations.

Finally, the usefulness of the models presented here are not limited to image segmentation. They can be applied to the processing of any spatially correlated multi-dimensional data, once a scan linking the variables under consideration has been established. There are numerous "multi-dimensional space-filling curves", which are scans mapping multi-dimensional space to one-dimensional domain [42]. For example, models considered in this paper would be directly applicable to the multi-dimensional data classification problem, once such a curve chosen. There are numerous "multi-dimensional space-filling curves", allowing mapping multi-dimensional space to one-dimensional domain [42]. For example, once such a curve - or scan - chosen, models considered in this paper are directly applicable to the unsupervised multi-dimensional non-stationary data classification problem.

**Funding:** This research was funded by the National Association of Research and Technology (ANRT), grant number 2018/0776.

**Acknowledgments:** The authors thank Julien FOUTH and Mokrane ABDICHE for their involvement in the project, as well as Segula Matra Automotive and the National Association of Research and Technology (ANRT) for the financial support.